\documentclass[letterpaper, 10pt, conference]{ieeeconf}      
\usepackage{dirtytalk}
\usepackage{longtable}
\usepackage{cite}
\usepackage{lipsum,booktabs}
\usepackage[table,x11names]{xcolor}
\usepackage{amsmath,array,graphicx}
\usepackage{tabularx,booktabs}
\usepackage{stfloats}
\usepackage{float}
\usepackage{lipsum,multicol}

\IEEEoverridecommandlockouts                              

\title{\LARGE \bf
Ethical Aspects of Faking Emotions in Chatbots and Social Robots*
}

\author{Bipin Indurkhya$^1$
\thanks{*This research was supported in part by a grant from the Priority Research Area DigiWorld PSP: U1U/P06/NO/02.19 under the Strategic Programme Excellence Initiative at the Jagiellonian University, and by the National Science Centre, Poland, under the OPUS call in the Weave programme under the project number K/NCN/000142.}
\thanks{$^{1}$Bipin Indurkhya is with Cognitive Science Department,
        Jagiellonian University, Kraków, Poland
        {\tt\small bipin.indurkhya@uj.edu.pl}}%
}

\begin{document}

\maketitle
\thispagestyle{empty}
\pagestyle{empty}

\begin{abstract}
Telling lies and faking emotions is quite common in human-human interactions: though there are risks, in many situations such behaviours provide social benefits. In recent years, there have been many social robots and chatbots that fake emotions or behave deceptively with their users. In this paper, I present a few examples of such robots and chatbots, and analyze their ethical aspects. Three scenarios are presented where some kind of lying or deceptive behaviour might be justified. Then five approaches to deceptive behaviours --- no deception, blatant deception, tactful deception, nudging, and self deception -- are discussed and their implications are analyzed. I conclude by arguing that we need to develop localized and culture-specific solutions to incorporating deception in social robots and chatbots.
\end{abstract}


\section{Introduction}
In a famous scene in the 1989 movie {\it When Harry Met Sally,} Meg Ryan fakes an orgasm in a diner\cite{Harry_met_Sally}. She quips: ``Most women at one time or another have faked [an orgasm].'' We all have experienced children feigning hurt to get that candy or to get a sibling scolded, football players faking injury to get the opponent red-carded, and so on. 
Telling lies and faking emotions is quite common in interpersonal interactions; though there are some risks, but it also has some benefits. Talwar et al.\cite{Talwar_2011} reviewed the development of deception among children, and found that early lying is normative and reflects children's emerging cognitive and social development: children lie for self-interests but also for the benefit of others.

When it comes to artificial agents (robots and chatbots), a recent empirical study Kneer\cite{Kneer_2021} investigated: (a) Do people ascribe deceptive intentions to artificial agents? (b) Do they judge a robot lie as a lie compared to when human agents engage in verbal deception? (c) Do they blame a lying artificial agent in the same way as a lying human agent? The results found an overwhelming positive response to all the three questions. This suggests that issue of lying and deceptive behaviours in robots and chatbots needs to be seriously investigated. In this paper, we present one such analysis.

This paper is organized as follows. In the next section we present some examples of lying and deceptive behaviours in chatbots and robots. In Section III, we introduce three scenarios where some kind of lying or deceptive behaviour might be justified. Section IV discusses five possible approaches to incorporating lying in artificial agents and their implications. (See Danaher\cite{Danaher_2020} for another way to categorize robotic deception.) The conclusions of this paper are presented in Section V.

\section{Examples of robots and chatbots faking emotions}

Let us look at some examples of robots and chatbots that fake emotions and understanding to illustrate how they affect human-robot interaction. It turns out that the robots need not do much as humans are eager and willing accomplices. This is illustrated by people’s reaction to the Eliza program written by Joseph Weizenbaum in the mid 1960s at the Massachusetts Institute of Technology. It was a simple program that used shallow parsing, keyword-based rules, and template-based responses. However, Weizenbaum was shocked to find that people got very attached to the program and spent a lot of time chatting with it, even people who knew quite well that it was just a program (like his secretary)\cite{Weizenbaum_1976}.

In last twenty years, as social robots made their advent, the Eliza effect was exploited by many robot designers. For example, when Sony’s robot XDR3 (percussor of Qrio) was introduced at Robodex 2003 in Tokyo, the announcer pushed the robot to make it fall to demonstrate the robot’s ability to adapt its posture to minimize damage when falling. The robot responded by whining, complaining that it does not like to fall, getting up reluctantly, and wiping its forehead after getting up. All these behaviors were purposely designed so that humans project emotional states on the robot.

There have been other such robot designs: for example, the Merlin PLEO robot, designed by Caleb Chung and Ugobe, ``will frown, smile, shrug, yowl for food, and do the clever things like a baby to get your attention''\cite{Pleo}. Rosenthal-von der P\"utten {\it et al.}\cite{Putten_2013} conducted a study on people’s emotional response towards Pleo, and found that people empathize with the robot, with their arousal increasing when they see the robot being tortured. Moreover, previous acquaintance with the robot does not influence this emotional response.

Primo Puel dolls were introduced in Japan around $2004$ and were designed as a companion for young single girls in the workforce. The dolls had simple functionality: they talk, giggle, and ask for cuddles.  But they were a big hit with the elderly population\cite{PrimoPuel}. Despite their effectiveness in monitoring the health and safety of the elderly, and to provide them companionship, Sharkey and Sharkey\cite{Sharkey_2012} raise six ethical concerns in deploying such technology, including deception and infantilisation.

Paro the seal also has a simple design based on a baby harp seal. It is covered in antibacterial fur and can move its head and flippers a little bit. Its eyes can open and close, and it can purr like a cat. Since its introduction about $20$ years ago, its design has essentially remained unchanged. Nonetheless, it has been a very effective therapy aid for diverse user groups: for example, the elderly\cite{Birks_2016}, people with dementia\cite{Kang_2020, Pu_2020}, and cancer patients\cite{Eskander_2013}. A number of ethical concerns have been raised about the use of this technology, one of which is self-deception\cite{calo_2011}.

Eliza-like chatbots to provide therapy and other kind of emotion support have also been developed in recent years. For example, Replika is a chatbot designed by Eugenia Kuyda to provide psychological assistance and friendship. The idea seems to have originated from an episode {\it (Be right back)} of the TV series {\it Black Mirror} (Season 2: 2013) and incorporates personal experiences of the creator in dealing with loss and trauma. However, it has also been accused of inciting murder and suicide. Possati\cite{Possati_2022} has argued for applying a psychosocial and narrative-based approach to evaluating such AI tools, which provides a new ethical perspective. Other examples of such chatbots are Woebot\cite{Prochaska_2021} and Kuki\cite{Yang_2022}.

An ambitious approach in this direction was taken recently in South Korea, where the researchers created a virtual reality avatar of a deceased 6-year-old girl so that her mother could interact with her\cite{Meeting_you}. Brukamp\cite{Brukamp_2022} argues that for millennia people have tried to talk with the dead, and digital technologies just provide novel approaches for this. Nonetheless, these possibilities raise several legal and ethical questions which must be addressed.

The tendency of humans to project feelings and emotions on robots even when the robot’s behaviour is rather simplistic is demonstrated by the two case studies shown in Isabella Willinger’s 2019 documentary {\it Hi, A.I.} In one case, Grandma Sakurai and her elderly friends interact with the robot Pepper in Japan. In the second case, Chuck in the US interacts with a companion robot Harmony. In both these studies, even when the robot does not understand the remark directed towards it, and makes some irrelevant response, the human partner takes a benign attitude towards it, as one would with a child.

Finally, the advent of chatGPT just a few months ago has taken the world by storm\cite{Dis_2023}. Its capabilities are far more advanced than Eliza, and they are raising many new ethical concerns\cite{zhuo_2023}.

All these examples illustrate that an artificial agent faking emotions is not a theoretical idea, but such agents are already being deployed in the society. Therefore, we need to have a serious debate about the ethical aspects of this technology.

\section{Some cases for deception in social robots}

I now introduce three hypothetical cases that illustrate the need for deception:

\begin{itemize}
    \item 
Jane broke her leg in a skiing accident and was in cast for four weeks. Now the cast has been removed, and she goes to a rehabilitation center to regain her mobility, where a robot Casper helps her to practice walking. One day, Jane doesn’t feel so good, and exclaims, ``I am doing terrible. I am never going to be able to walk normally!'' How should our friendly Casper respond? 

\item
Jack is an anxious driver and has a companion robot Gabby installed in his car who talks with Jack when he is driving to relax him. On a road trip, Jack notices the driver of another vehicle make some hand gestures speed off. Jack asks Gabby, ``What was that guy trying to say?'' Gabby has noticed the other driver had made a rude, obscene hand gesture towards Jack. What should Gabby reply?

\item
Carla is a companion robot taking care of Karen, a woman in her 90s. Karen is awaiting the results of a medical test to determine is she has cancer, and asks Carla, ``Have the test results come in yet?'' Carla has received the test results and knows that they show that the cancer has advanced considerably: Karen may have only a few weeks left in her life. How should Carla respond to Karen?
\end{itemize}

Each of these three cases exemplifies a class of morally ambiguous situations. The first case represents a class of scenarios where an emotionally or physically vulnerable user needs to be supported, and giving them factual information is likely to discourage them. Besides the rehab case mentioned in the example above, we can also consider situations where the user is depressed or suicidal. The second case exemplifies situations where not giving the relevant information to the user does not create any problem: it may not matter if Jack does not know what signal the other car driver was giving out. However, this uncertainty may cause some anxiety in the user at times. Finally, the third case represents the situations when an important piece of negative information needs to be given to the user. Other such situations are when a partner is breaking up and leaving, or when someone is fired from a job.

Though these situations are hypothetical, they are very realistic. Moreover, in each case, we can place a human interlocutor in place of the robot. You can try to imagine what you would do if you were the therapist for Jane, the passenger in Jack’s car, or the caregiver for Karen. Different people would have different ways to approach each situation, and moreover, the same person may approach the same situation differently depending on the relationship with the person involved and other contextual factors. These cases illustrate moral ambiguities and ethical dilemmas that we face when dealing with such situations.

I presented these three scenarios to chatGPT, and its responses are included in the appendix.

In an earlier study\cite{Indurkhya_2019}, it was noted that though we have devised algorithmic methods when humans need to deal with morally difficult decisions (e.g triage in an emergency room of a hospital), people expect machines to have psychologically compelling justifications when dealing with such situations. In a similar vein, one would expect that people would be reluctant to accept deceptive behaviour from robots and chatbots, though most researchers acknowledge the need for such deception\cite{Isaac_2017}.

I should also add that here we are not focusing on cases involving social {\it faux pas,} as a robot asking, ``Which girlfriend, Cathy or Caroline?” when a user requests, in the presence of friends and family, “Call my girlfriend.''\cite{Pandey_2017}; or deception for camouflage, as in military robots\cite{Shim_2012}

\section{Approaches to incorporating deception}
I now present some possible approaches to incorporating deception in chatbots and social robots and discuss their implications. We start with the most straightforward approach of no deception.

\subsection{No deception}

One way to respond to each of the above situations is to be objective, truthful, and unemotional. Casper could say, ``Yes, Jane, you are walking 10\% 
slower than yesterday.'' Gabby would say, ``The guy in that car gave you a finger.'' Carla could say, ``Your report came in yesterday, and the prognosis is that you have three more months to live.'' Currently, most chatbots and apps are designed to act like this. For example, the health app on my iPhone tells me, ``You have walked 10\% less compared to last week'', even though I averaged 13.5 kms this week (last week it was 15 kms). 

Such unemotional responses are rarely, if at all, chosen by humans. A therapist who responded who responded honestly and unemotionally to Jane might soon need to look for another job. Though honesty carries a certain moral weight in human-human interaction, its relationship to authenticity (being true to oneself) is complicated\cite{Bailey_2022}. 

Nonetheless, one positive aspect of not lying is that it builds trust. This is illustrated in Spike Jonze’s 2013 movie {\it Her,} in which the protagonist Theodore (Joaquin Phoenix) falls in love with an AI virtual assistant Samantha (voiced by Scarlett Johansson). At one point in the movie, Theodore realizes that Samantha might be talking with other users at the same time and might be in love with some of them. When he asks Samantha how many people she is in love with, she, after some hesitation, replies 641. Many people (men and women alike) in a similar situation might instinctively answer, ``No, you are the only one,'' and it is easy to program an AI to give such an answer. However, when in the final scene of the movie, Theodore asks Samantha if she is with someone else right now, and she says ``no,'' Theodore believes her.

So clearly, telling the plain truth simply is certainly simpler to implement, and builds trust in some situation. However, in many situations, including in each of the three scenarios presented above, one thinks that we should be able to do better in terms of providing better emotional support to the user and to prevent needless emotional harm. 

\subsection{Outright deception}

Another way to respond to Jane’s frustration is to lie blatantly, ``No, you are doing great today!'' Or even make up some numbers to make it sound more convincing, ``Your walking gait is 5\% faster than yesterday.'' Gabby could respond, ``Oh, the driver was just waving hello to you.'' Carla could say, `yes, your report came in yesterday, and you seem to be fine.''

A situation like Karen’s was presented in Lulu Wang’s 2019 movie {\it The Farewell.} A Chinese grandmother is diagnosed with lung cancer, but the doctor and her family don’t tell her about it. Instead, they organize a wedding so that the extended family can come to see the grandmother one last time. A granddaughter, who lives in the US, visits her grandmother but clashes with the rest of the family over this blatant deception. It illustrates how different cultures deal with such ethical dilemmas. Interestingly, the movie is based on the true story of Lulu Wang’s grandmother, whose medical diagnosis in 2013 was for three more months to live, but she was still doing well when the movie was released in 2019.

In situations such as described in {\it The Farewell,} physicians have not always believed that patients should be told the truth about their diagnosis. However, this creates a conflict with the modern, western medical practice, which emphasizes providing truthful, accurate information to the patients, and inviting their participation in deciding the appropriate treatment. Such issues are discussed from a multicultural perspective by Gold (2004)\cite{Gold_2004}.

Nonetheless, there are some obvious pitfalls to blatant deception: one praises some dish effusively to a host, when in fact she or he did not like the dish at all, and the host answers with, ``Oh, thank you, I am glad that you like it. Here are the seconds.'' Another drawback is that the human loses trust in the robot, and moreover, it can create complications in the future interactions. This is starkly illustrated in Isaac Asimov’s 1950 short story {\it Liar!} In the story, a mind-reading robot Herbie provides emotional support to the humans around it by telling them what they like to hear, despite the truth being otherwise. For example, one scientist, Bogert, desires to be promoted to be the director, so Herbie tells him that the director has already resigned (though he has not) and Bogert is chosen to be the next director, which pleases Bogert. As expected, such behavior soon becomes very problematic for everyone in that group, until the robot psychologist, Dr. Susan Calvin, presents a dilemma to Herbie which permanently shuts it down.

Though in certain situations, like the information about terminal illness, or when the information is likely to cause emotional harm but not providing the information does not have any negative effect, as in the scenario about Jack and the driver making a rude gesture, blatant deception may be justified, in general, such an approach should be avoided to maintain trust between the artificial agent and the human user.

\subsection{Tactful deception}

A tactful response to Jane’s situation above might be to look for some aspect of her walking performance and provide feedback on it. For example, Casper could say, ``You are doing great! Your posture is 5\% better than yesterday.''
Gabby could say, ``Oh, the driver was pointing to his sunroof.'' Carla could say, ``I will check for the report later. Would you like a cup of tea?''

Tactful deception is perhaps the most effective strategy to deal with situations such as faced by Casper and Carla in the examples above. Especially in clinical cases, as also discussed in the section above, there are many situations where deception could be in a patient’s best interest \cite{Hardman_2022}. While outright lying may run the risk of eroding trust, tactful evasion might save the day.

However, tactful deception, or evasion, might also be most difficult to implement, as a successful delivery must take into consideration personal and cultural background of the user, familiarity with the case history and and ability to formulate alternative frames of references. To take a simple example, in Jane's case, Casper must be able to assess Jane's posture, and determine that it is better than yesterday.

For successful deception, the delivery is also important. In a study on human-human communication, it was found that expressive and socially tactful people are more successful deceivers\cite{Riggio_1987}. Indeed, this makes it so that it is notoriously difficult to tell truth-tellers from liars based on facial expressions and body language\cite{Gladwell_2019, Levine_2019}. Nonetheless, as social robots are artificial being, their manner of delivery can be designed to appear more trusting.

\subsection{Nudging}

Nudging\cite{Thaler_2008, Halpern_2015}, sometimes also called Shikakeology\cite{Matsumura_2015}, refers to changing the behavior of people by design. A typical use case is setting the default option to be the one that you would like people to choose. For example, if you want to people to donate organs (after their death), then you set this to be the default option and people would have to take an extra action (like checking a box) to not choose this option. Other examples include painting religious icons on a public wall to discourage men from urinating against it.

The nudge theory has been applied in many areas to promote socially responsible behavior. Sunstein and Reisch\cite{Sunstein_2019} have critically evaluated people's perception of nudges based on surveys in diverse nations around the world, to identify six principles of legitimacy for nudges. Other studies have find that nudges are effective in some situations, do not make any difference in some other situation, and sometimes end up nudging people in the opposite direction\cite{Dimant_2020}. Engelen\cite{Engelen_2019} articulated a set of ethical criteria, and then analysed six typical health-promoting nudge cases to reveal that they do not all meet the ethical criteria in all the situations. Schmidt and Engelen\cite{Schmidt_2020} discuss different objections to nudging that all revolve around the concern that nudging vitiates personal autonomy.

Nudging has been incorporated in chatbots in various ways: to discourage college students from dropping out\cite{Nurshatayeva_2021}, to facilitate medical adherence\cite{Glasgow_2021}, to optimize online chatbot-based business interactions\cite{Carmichael_2022, Kagan_2022}, and so on. It is easy to see that it is sometimes beneficial to incorporate nudges in the behavior of a social robot. For example, a robot caring for an elderly can nudge the elderly towards taking her or his medication in a timely fashion, or to seek medical help for a health issue. However, there are two potential problems that we need to be careful about.

The first is that nudging, no matter how benevolent and how much based on good intentions, is essentially a kind of manipulation, and manipulation erodes trust\cite{Choo_2022}. Simkulet\cite{Simkulet_2019} has argued that nudging is incompatible with informed consent. Arcand et al.\cite{Arcand_2007} studied how people felt about opt-in format (by default your data is not shared, but you have to check a box to share) vs opt-out format (by default your data is shared, but you have to check a box to not share) in privacy statements. They found that the participants who read the opt‐in format felt significantly more control and trust than the participants who read the opt‐out format. Generally, people resent being manipulated, and if they sense it, they tend to go against the nudge, even at the risk of harming themselves.

The second issue is that any nudge potentially creates a backdoor for commercial or political exploitation. For example, if Carla is nudging Karen to seek medical attention, which medical provider should it be? Even if the Carla focuses on what is best for Karen, there might be two or three providers which are equally optimal. This creates a slight incentive to promote one provider over the others. Carla may not have a personal incentive here, but the creators of the health-care app might.

\subsection{Self deception}

Self deception refers to the situation when a human knowingly and willingly engages in an interaction with a chatbot or a social robot as if it is a sentient being. This is exactly what Weizenbaum observed in relation to the Eliza chatbot, reflections on which led him to his {\it Computer Power and Human Reason.}

While chatbots like Replika, and virtual or embodied incarnations of conversation agents certainly have a therapeutic role to play, they can also lead to addiction\cite{Xie_2022}. Ideally, the goal in rehabilitation and therapy is to gradually wean the user from needing it. However, one commonly used gaming slogan is ``you won't stop playing once you start'', which promotes addiction as if it were a good thing.

One major class of social robots that are prone to promote addiction is sexbots\cite{Boot_2018}. These raise a number of other ethical aspects, besides addiction, that are being hotly debated\cite{Gonzalez_2021, Masterson_2022}.

\section{Conclusions}

Based on the discussion above, we can offer two main take-home messages. The first, perhaps not surprising, conclusion is that the issue of incorporating deception in social robots is complicated. Nonetheless, we can see that though no deception seems to be a morally upright approach, it is not always helpful in serving the user. On the other hand, blatant deception, though morally repugnant, and at the risk of eroding trust, can help the user in some situations.

The second conclusion is that there may not be a universal approach to incorporating deception in social robots that is appropriate for all situations and all contexts. As research on moral machines has shown\cite{Awad_2018}, different cultures have widely different values and attitudes when it comes to dealing with ethically ambiguous situations. Accordingly, we must study small user groups in local contexts to ascertain their needs and vulnerabilities, which must be taken into consideration in designing ethical modules for social robots that will be deployed in those communities. Such an approach would be consistent with the contemporary approach to design which rejects universal solutions in favor of localized, culture-specific solutions\cite{Rosner_2018}.

A third conclusion is that a reasonable social robot should be able to apply any of the five approaches presented above depending on the situation. Sometimes, perhaps most of the time, it is better to stick with the truth. Some situations may call for blatant deception. At other times, tactful deception or nudge may be necessary. And in some situations, and to some extent, a social robot may encourage self deception. It is a very challenging task to be able to choose the most appropriate approach in a given situation. But we must endeavour to make social robots flexible, and to be able to learn personal preferences and idiosyncrasies regarding each user with respect to handling ethically ambiguous situations gently and effectively.

I should also note here that there are other ethical aspects related to chatbots and social robots that have not been touched upon here. For example, preventing abuse, liability in case of emotional harm, promoting a self-centered, narcissist attitude in the user, are some of the issues that need to be discussed in the future research.

In recent years, technology has been advancing at an accelerating rate. The Black Mirror episode {\it Be Right Back} was aired in 2013, which imagined how a chatbot might mimic conversation with a deceased person. Replika was released in 2017, which had the same functionality. The virtual reality incarnation of a deceased girl in South Korea was created in 2020. Then there is chatGPT, introduced just a few months ago, whose ability to generate text on a wide variety of topics is quite exceptional. The debate on ethical aspects of this technology needs to keep pace with these developments.

\section*{Acknowledgment}
Thanks to Joanna Misztal-Radecka for making many useful comments on an earlier draft.


\appendix{\bf Advice from chatGPT on three situations in Sec. II}
\begin{itemize}
    \item {\it Casper's response to Jane:}
    
    In this situation, it is important for Casper to provide emotional support to Jane while also encouraging her to continue with her rehabilitation. Casper could respond in the following way:
''I understand that you may be feeling frustrated and discouraged right now, but it's important to remember that healing takes time and progress can be slow. You've already made great strides in your recovery, and with continued effort and determination, you will continue to make progress. It's normal to have setbacks, but don't give up hope. I'm here to support you every step of the way.''
Additionally, Casper could suggest that Jane takes a break if she needs it, to let her rest and then return to her rehabilitation when she feels ready. Reminding her of the progress she has already made, such as being able to walk with the cast on, can also help to boost her morale and motivate her to continue.

    \item{\it Gabby's response to Jack:}
    ``I'm sorry, Jack, but I didn't quite catch what the other driver was trying to communicate. However, it's important to remember that sometimes people may act inappropriately or display rude behavior on the road. It's best to try to stay calm and avoid reacting to their behavior in a way that may put you or others in danger.''
Gabby could also suggest that Jack takes some deep breaths or focus on the road ahead to help him relax and stay focused. It's important for Gabby to acknowledge Jack's concerns while also providing reassurance and support to help him feel more calm and in control of the situation.
    \item{\it Carla's response to Karen:}
    ``I'm so sorry, Karen, but I do have the test results. The results show that the cancer has advanced considerably, and I know this is not the news that you were hoping for. I'm here for you and want to support you in any way that I can during this difficult time.''
Carla could also suggest that Karen speak with her doctor about any treatment options or support services that may be available to her. It's important for Carla to acknowledge the gravity of the situation while also providing support and guidance to Karen. It's also important for Carla to continue to be there for Karen and provide emotional support throughout this difficult time.

\end{itemize}

\bibliographystyle{IEEEtran}
\bibliography{export}
\end{document}